\documentclass[10pt, a4paper]{article}
\usepackage{lrec2022} 
\usepackage{multibib}
\newcites{languageresource}{Language Resources}
\usepackage{graphicx}
\usepackage{tabularx}
\usepackage{soul}
\usepackage{subcaption}
\usepackage{multirow}
\usepackage{float}
\usepackage{xcolor}
\usepackage{booktabs}
\usepackage{colortbl}
\usepackage{amssymb}
\usepackage{xcolor}
\usepackage{xspace}
\usepackage{tikz}
\usepackage{pifont}
\usepackage{hyperref}
\usepackage{xstring}
\usepackage{color}
\usepackage{epstopdf}
\usepackage[utf8]{inputenc}

\usepackage{titlesec}
\titleformat{\section}{\normalfont\large\bfseries\center}{\thesection.}{1em}{}
\titleformat{\subsection}{\normalfont\SmallTitleFont\bfseries\raggedright}{\thesubsection.}{1em}{}
\titleformat{\subsubsection}{\normalfont\normalsize\bfseries\raggedright}{\thesubsubsection.}{1em}{}
\renewcommand\thesection{\arabic{section}}
\renewcommand\thesubsection{\thesection.\arabic{subsection}}
\renewcommand\thesubsubsection{\thesubsection.\arabic{subsubsection}}

\newcommand{\cmark}{\ding{51}}%
\newcommand{\xmark}{\ding{55}}%

\newcommand{\correctmark}{{\color{green}\cmark}\xspace}
\newcommand{\wrongmark}{{\color{red}\xmark}\xspace}
\newcommand{\lightrule}{\arrayrulecolor{lightgray}\midrule\arrayrulecolor{black}}
\definecolor{amber}{rgb}{1.0, 0.75, 0.0}

\usetikzlibrary{shapes.geometric}
\newcommand{\stars}[2][fill=amber,draw=white]{\begin{tikzpicture}[baseline=-0.35em,#1]
\foreach \X in {1,...,5}
{\pgfmathsetmacro{\xfill}{min(1,max(1+#2-\X,0))}
\path (\X*0.9em,0) 
node[star,draw,star point height=0.20em,minimum size=0.8em,inner sep=0pt,
path picture={\fill (path picture bounding box.south west) 
rectangle  ([xshift=\xfill*1em]path picture bounding box.north west);}]{};
}
\end{tikzpicture}}

\newcommand{\correctanswer}[1]{\correctmark #1}
\newcommand{\wronganswer}[1]{\wrongmark #1}
\newcommand{\copainstance}[3]{\multicolumn{2}{l}{#1} \\ \multicolumn{2}{l}{#2 \quad #3}}
\newcommand{\copainstanceshort}[2]{\multicolumn{2}{l}{#1 \correctanswer{#2}}}
\newcommand{\example}[2]{#1 \\ \lightrule #2}
\newcommand{\ratedexplanation}[2]{\stars{#1} \hspace{-1.1em} & #2 \\}
\newcommand{\triple}[3]{(\textit{``#1''}, \texttt{#2}, \textit{``#3''})}

\newcommand{\crowdinput}[3]{\textit{\underline{~~#1~~}}~\texttt{#2}~\textit{\underline{~~#3~~}}}

\title{COPA-SSE: Semi-structured Explanations for Commonsense Reasoning}

\name{Ana Brassard,\textsuperscript{1, 2} Benjamin Heinzerling,\textsuperscript{1, 2} Pride Kavumba,\textsuperscript{2, 1} Kentaro Inui\textsuperscript{2, 1}} 

\address{\textsuperscript{1}Riken AIP, \textsuperscript{2}Tohoku NLP Lab \\
         \{ana.brassard, benjamin.heinzerling\}@riken.jp, \\ 
         kavumba.pride.q2@dc.tohoku.ac.jp, inui@tohoku.ac.jp\\}

\abstract{
We present \oursfull{} (\ours), a new crowdsourced dataset of 9,747 semi-structured, English common sense explanations for \copafull{} (\copa) questions.
The explanations are formatted as a set of triple-like common sense statements with \cpnet{} relations but freely written concepts.
This semi-structured format strikes a balance between the high quality but low coverage of structured data and the lower quality but high coverage of free-form crowdsourcing.
Each explanation also includes a set of human-given quality ratings.
With their familiar format, the explanations are geared towards commonsense reasoners operating on knowledge graphs and serve as a starting point for ongoing work on improving such systems. The dataset is available at \href{https://github.com/a-brassard/copa-sse}{https://github.com/a-brassard/copa-sse}.
\newline 
}

\begin{document}

\def\ours{COPA-SSE}
\def\oursfull{Semi-Structured Explanations for COPA} 
\def\bcopa{Balanced COPA}
\def\copa{COPA}
\def\csqa{CommonsenseQA}
\def\cose{CoS-E}
\def\cpnet{ConceptNet}
\def\copafull{Choice of Plausible Alternatives}

\maketitleabstract

\section{Introduction}
\label{sec:intro}
While there are many datasets for question answering and commonsense reasoning \cite{rogers2021qa}, models are known to exploit shortcuts such as superficial cues in these datasets, which leads to artificially high evaluation scores \cite{gururangan2018annotation}.
One way to ensure models are reasoning as intended is to require explanations for their predictions \cite{bowman2021nlpbroken}.
A prominent example of such a setting is the Commonsense Explanations Dataset (\cose)~\cite{rajani2019explain}, which provides crowdsourced justifications of the correct answers expressed in free text.
While free-form crowdsourcing allows representing natural and diverse human reasoning, quality control is notoriously difficult \cite{daniel2018quality}.
At the other end of the spectrum are explanations that are fully grounded in a knowledge graph (KG), i.e., each element of the explanation corresponds to a node or edge in a KG. 
However, this structured approach is limited by the coverage of the KG, i.e., the explanation will be sub-optimal or impossible when the situation to explain is not covered by the KG.
Here, we adopt a semi-structured approach aiming to combine the best of both worlds---the coverage potential of open-ended crowdsourcing and quality control of structured data.

\begin{figure}[H]
\centering
\begin{subfigure}{\linewidth}
    \includegraphics[width=\textwidth]{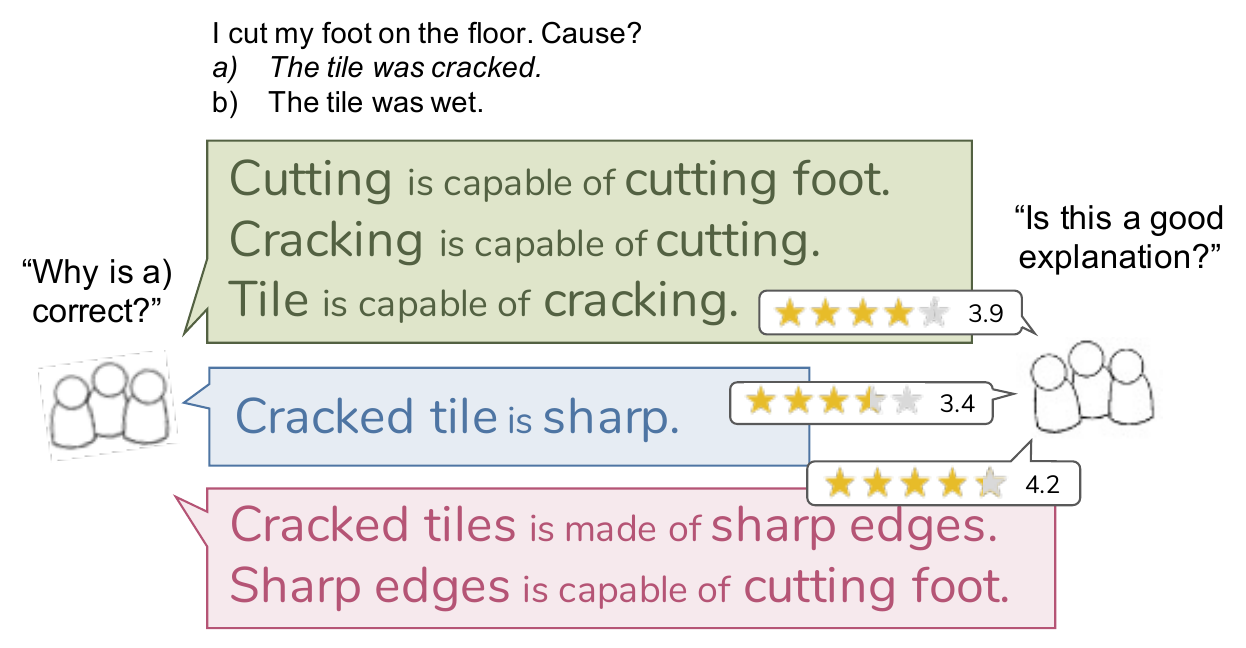}
    \caption{Collecting and rating explanations for \bcopa.}
    \label{fig:fig1-collection}
\end{subfigure}
\hfill
\begin{subfigure}{0.8\linewidth}
    \vspace{5px}
    \includegraphics[width=\textwidth]{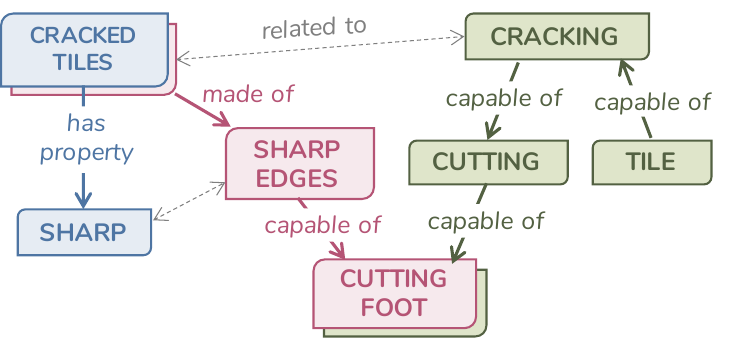}
    \caption{The explanations in graph form.}
    \label{fig:fig1-aggregation}
\end{subfigure}
\caption{Constructing \ours. Crowdworkers gave one or more triple-like statements explaining the correct answer which were then rated by different workers (a). Each statement consists of head and tail text linked by a \cpnet{} relation. The statements can be aggregated into an explanation graph (b) (\S \ref{sec:aggregation}).} \label{fig:fig1}
\label{fig:figures}
\end{figure}

Specifically, we introduce \oursfull{} (\ours), a new explanation dataset for the Choice of Plausible Alternatives (\copa{}) dataset
\citelanguageresource{roemmele2011choice}.\footnote{
    We use \bcopa{} \citelanguageresource{kavumba2019choosing}, a superset of \copa{}.}
Each explanation consists of a set of English statements, which, in turn, consist of a head text, a selected predicate, and tail text, mimicking \cpnet{} \citelanguageresource{speer2017conceptnet} triples (Figure \ref{fig:fig1}).
The head and tail texts are free-form, allowing an open concept inventory.
Each explanation{\parfillskip=0pt\par}
also includes quality ratings.
Note that \ours{} is not meant to extend existing commonsense knowledge graphs, but rather to be used as examples of extraction and/or generation \emph{results} based on a specific prompt (question).

In this paper, we introduce \ours{} (\S \ref{sec:our-dataset}), detail its construction (\S \ref{sec:crowdsourcing}), demonstrate a simple application (\S \ref{sec:experiments}), 
and discuss future use cases (\S \ref{sec:discussion}).
\ours{} is available for download at \href{https://github.com/a-brassard/copa-sse}{https://github.com/a-brassard/copa-sse}.

\section{\oursfull}
\label{sec:our-dataset}
\begin{figure}
  \centering
  \includegraphics[width=\linewidth]{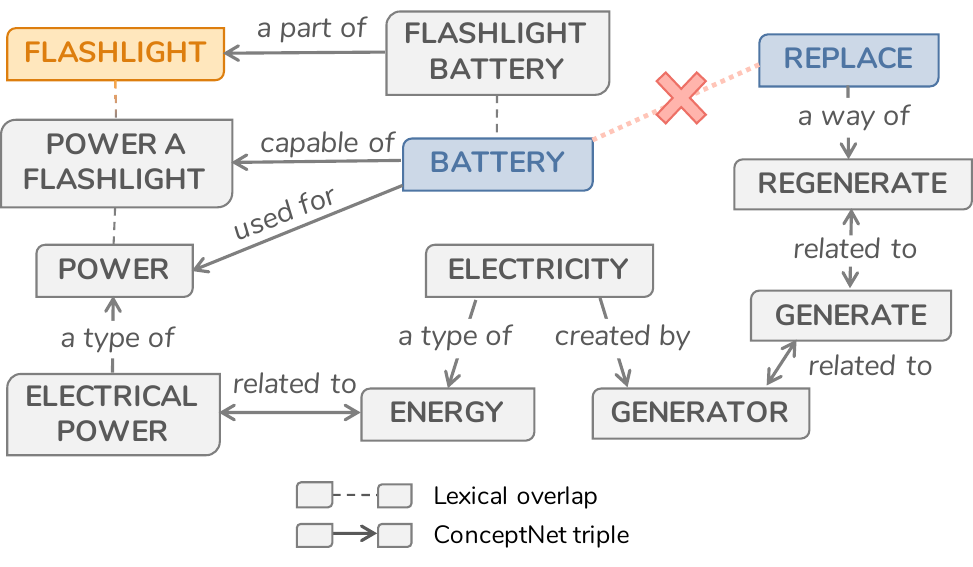}
  \caption{A manually extracted \cpnet{} subgraph to illustrate the caveats of only using existing resources. One author attempted to find paths connecting concepts from the question \textit{The flashlight was dead. Effect?} and the answer \textit{I replaced the batteries.} and was unable to find a meaningful path between \textit{battery} and \textit{replace.} The two concepts are connected but the path contains irrelevant facts to the point of being meaningless.}  \label{fig:manual-subgraph}
\end{figure}

\paragraph{Design goals.} Our goal is to add high-quality explanations to \bcopa.
Since the nature of a good explanation is subject of debate \cite{miller2019explanation}, we adopt a working definition: A good explanation is a minimal set of relevant common sense statements that coherently connect the question and the answer. 
For example, the fact \emph{Opening credits play before a film.}\ connects the question \emph{The opening credits finished playing.\ What happened as a result?}\ and its answer \emph{The film began.}
Commonsense KGs such as \cpnet{} provide such statements but have limited coverage \cite{hwang2021cometatomic20}.
For example, even if question and answer concepts are found in the KG, the paths between them can degenerate into long chains of statements that are neither minimal nor relevant (Figure~\ref{fig:manual-subgraph}).

In contrast to structured approaches, unstructured free-form text is not limited by KG coverage.
Previous work has elicited such free-form explanations from crowdworkers, but suffers from low quality.
For example, in a preliminary manual inspection of a random sample of 1,200 \cose{} explanations, one of the authors judged only a small fraction to be acceptable explanations in terms of relevance and thoroughness.

Aiming for a golden middle, we devise a semi-structured explanation scheme comprising a set of triple-like statements. 
Each statement consists of open-ended head text and tail text connected with a \cpnet{} relation.
In practice, crowdworkers created explanations by selecting a predicate from a list while providing free text for the two concept slots.
This format encouraged workers to provide explanations close to our definition without being restricted to a pre-defined inventory of concepts.
We refer to this combination of free text and \cpnet{} predicates as \emph{semi-structured explanations}.

Among related prior work, our approach is most similar to the explanation graphs by \cite{saha2021explagraphs}, which also combine ConceptNet relations and free text concepts, but differ in task, domain, and crowdsourcing protocol.

\renewcommand{\arraystretch}{1.1}
\begin{table*}[h]
\centering
\small
\sffamily 
	\begin{tabular}{lp{0.7\linewidth}}
        \toprule
	\example{
		\copainstance{
		The documents were loose. Effect?}{
		\correctanswer{I paper clipped them together.}}{
		\wronganswer{I kept them in a secure place.}
		} 
	}{
		\ratedexplanation{4.20}{Paper clip is used for loose documents.}
		\ratedexplanation{4.00}{Paper clips is used for keeping documents together. Paper clipping can be done to have the documents together.} 
		\ratedexplanation{3.70}{Paper clip is used for clipping paper together.} 
		\ratedexplanation{3.40}{Paper clip is used for organizing papers.} 
		\ratedexplanation{3.40}{Paper clip can be done to keep papers together.} 
		\ratedexplanation{3.20}{The paper clipped is a way of holding the papers together.} 
	}
	\midrule

	\example{
		\copainstance{
			They lost the game. Cause?}{
			\wronganswer{Their coach pumped them up.}}{
			\correctanswer{Their best player was injured.}
		}
	}{
		\ratedexplanation{4.2}{Game is a team work. Player is a part of a team. Player injured causes team not working properly. Team not working properly causes lose the game.} 
		\ratedexplanation{4.0}{Best player is a part of the team. Injury of the best player causes the team to lose.} 
		\ratedexplanation{3.67}{Their best player being injured causes the team to lose.} 
		\ratedexplanation{3.56}{Teams is made of players. Injuries is capable of causing losses.} 
		\ratedexplanation{3.5}{Injury is capable of causing loss.} 
		\ratedexplanation{2.22}{The team causes the injury.} 
	  }
          \bottomrule                    
\end{tabular}

\caption{Examples of collected and rated explanations for \bcopa~questions.}
\label{tab:more-examples}
\end{table*}

\renewcommand{\arraystretch}{0.95}

\begin{figure}
  \centering
  \includegraphics[width=\linewidth]{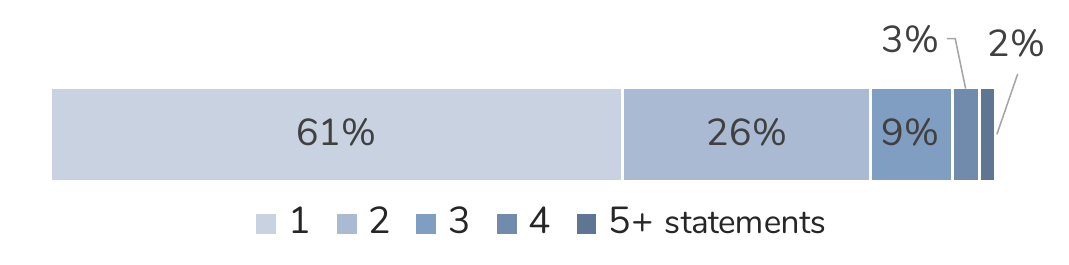}
  \caption{Number of statements per explanation.} 
  \label{fig:num-statements}
\end{figure}

\begin{figure}
  \centering
  \includegraphics[width=\linewidth]{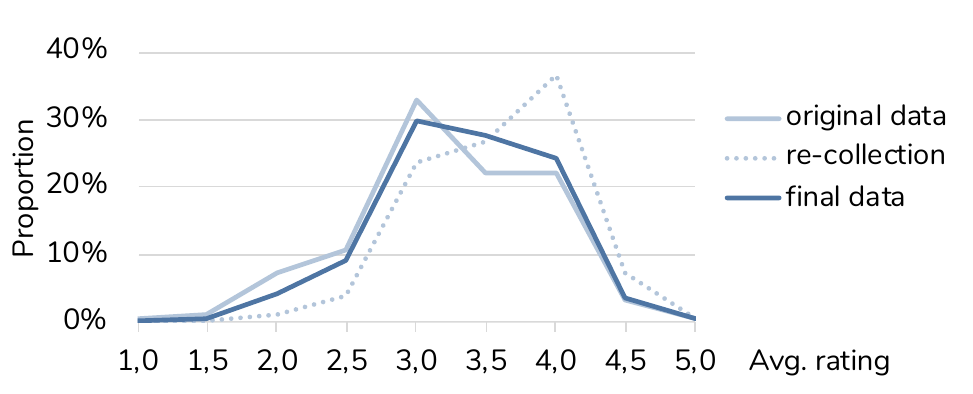}
  \caption{Average rating distribution before (original data) and after the re-collection round (final data). Values are rounded to the nearest half-star.}
  \label{fig:distribution}
\end{figure}

\paragraph{Dataset statistics.} Table \ref{tab:more-examples} shows examples of \ours{} explanations. \ours{} contains $9,747$ commonsense explanations for $1,500$ \bcopa{} questions.
Each question has up to nine explanations given by different crowdworkers.
We provide the triple-format described above, as well as a natural language version obtained by replacing \cpnet{} relations with more human-readable descriptions.
$61\%$ of explanations are only one statement while the other $39\%$ comprise two or more, with the longest explanation being ten statements (Figure~\ref{fig:num-statements}).
Each explanation has a quality rating on a scale of $1$ to $5$ as given by crowdworkers.
Figure~\ref{fig:distribution} shows the rating distribution after initial collection (original data).
To guarantee that each \bcopa{} instance is explained by high-quality explanations, we collected additional explanations until most \bcopa{} instances ($98\%$) had at least one explanation rated 3.5 or higher (final data).
In other words, $98\%$ of the questions have at least one highly-rated explanation. 
Initially, $38\%$ of all explanation were over this threshold, which increased to $44\%$ after the additional collection run.
We kept the lower-quality explanations as they can be useful negative samples, e.g., in contrastive learning settings where they can act as sub-optimal examples in terms of thoroughness or relevance.
Finally, we created additional aggregated versions of the explanations by merging or connecting similar concepts. 
We now describe each step in more detail.

\section{Crowdsourcing Protocol}
\label{sec:crowdsourcing}
Crowdworkers were asked to provide one or more statements that connect the question and the answer in a triple format: a free-form head text, a selection of \cpnet{} relations, and a free-form tail text, together forming a commonsense statement (\S\ref{sec:collection}).
Each set of statements was then rated by five different workers (\S\ref{sec:rating}).
To gather more high-quality explanations, we invited workers whose explanations were highly rated to provide additional explanations (\S\ref{sec:recollecting}).
Section~\ref{sec:qualifs}{} lists worker qualifications and compensation, followed by further post-processing in Section~\ref{sec:aggregation}.

\begin{figure*}[h]
  \centering
  \includegraphics[width=0.9\linewidth]{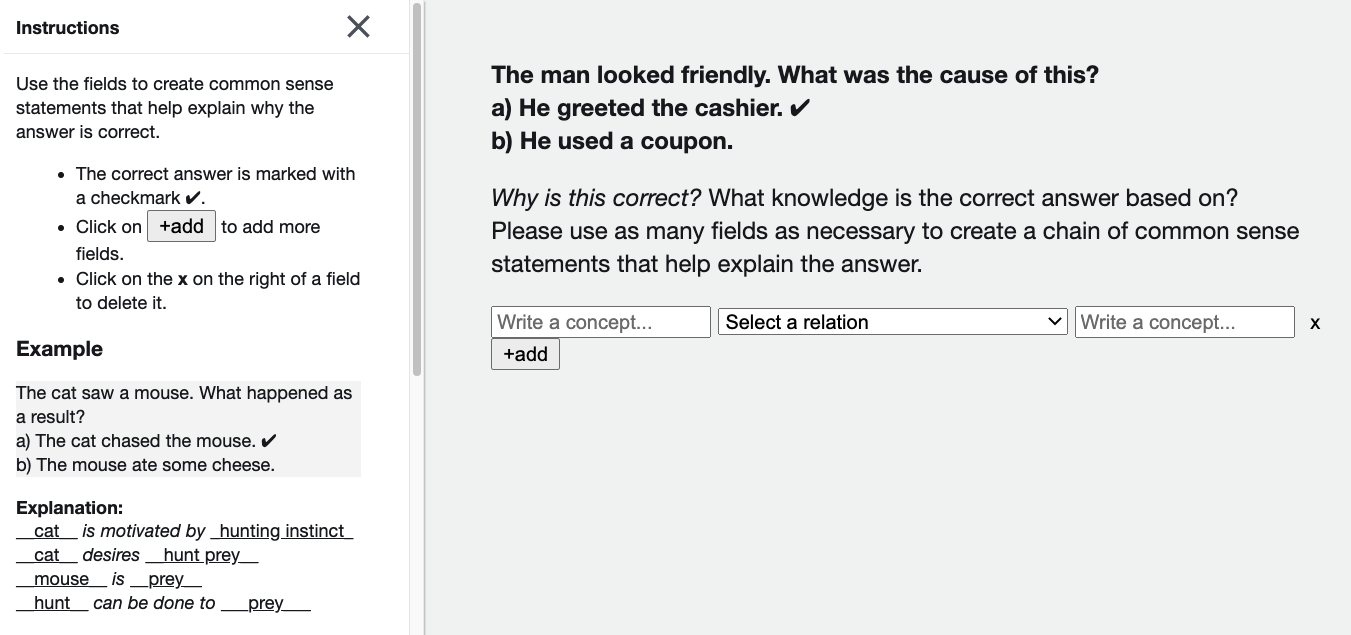}
  \caption{Form for collecting explanations.} 
  \label{fig:collection}
\end{figure*}

\begin{figure*}[h]
  \centering
  \includegraphics[width=0.9\linewidth]{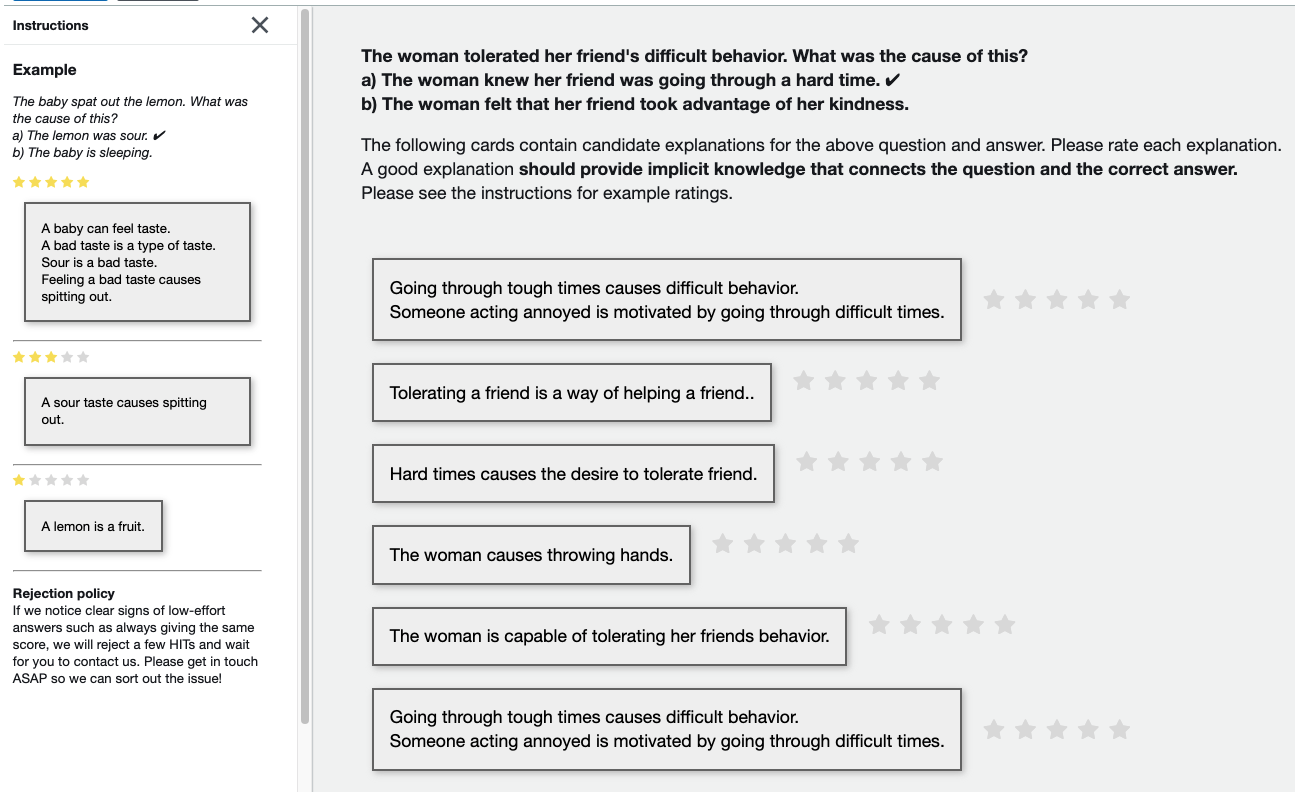}
  \caption{Form for rating Explanation.} 
  \label{fig:rating}
\end{figure*}

\subsection{Collecting Explanations}
\label{sec:collection}

Figure \ref{fig:collection} shows our collection form.
Workers were given a \copa{} question and two answer choices with the correct one marked. 
The input row below consists of two text fields for inputting concepts and a drop-down box for selecting the relation between them.
Workers could increase the number of rows to provide explanations with multiple statements, as they were encouraged (but not forced) to do.
The relations are a subset of \cpnet{} predicates which we selected and translated into human-readable English for easier understanding by non-experts.\footnote{E.g., \texttt{A HasSubevent B} is shown as \texttt{A} \textit{happens during} \texttt{B}. The text-form explanations retained the original surface form, while in the triple format they are changed back to match \cpnet.} 
For example, the input \crowdinput{an apple}{is a}{fruit} corresponds to the statement \textit{``An apple is a fruit.''} and the triple \triple{an apple}{IsA}{fruit}.
	
Free-form text guarantees neither consistent granularity nor chains of statements connected by matching concepts.
For example, a phrase such as \textit{``the act of eating a sweet fruit''} can be given as tail text, even though the next statement might not include that same phrase. 
We opted to leave this freedom as longer statements can still form coherent explanations, and, as we found in preliminary runs, introducing strict constraints might lead to unnatural and/or less informative explanations.
Overly long statements were rare, as most workers followed the simple examples we provided. 

\renewcommand{\arraystretch}{1.1}
\begin{table*}[h]
\centering
\small
\sffamily
	\begin{tabular}{lp{0.8\linewidth}}
        \toprule

		\example{
			\copainstanceshort{The woman sensed a pleasant smell. Effect?}{She was reminded of her childhood.}
		}{
			\ratedexplanation{5.0}{Pleasant smell is a way of bring happiness. Happiness causes nostalgia. Nostalgia is related to a smell. Smell causes her to think her childhood.}
		}
		\midrule

		\example{
			\copainstanceshort{The flashlight was dead. Effect?}{I replaced the batteries.}
		}{
			\ratedexplanation{5.0}{Batteries is used for flashlights. Power is created by batteries. Replacing batteries is a way of restoring power.}
		}
		\midrule

		\example{
			\copainstanceshort{The car looked filthy. Effect?}{The owner took it to the car wash.}
		}{
			\ratedexplanation{5.0}{The owner desires clean car. Car wash is used for washing cars.}
		}
		\midrule

		\example{
			\copainstanceshort{My favorite song came on the radio. Effect?}{I sang along to it.}
		}{
			\ratedexplanation{1.0}{This is a symbol of simple.}
		}
		\midrule

		\example{
			\copainstanceshort{The rain subsided. Effect?}{I went for a walk.}
		}{
			\ratedexplanation{1.0}{The rain has a fresh smell.}
		}
		\midrule

		\example{
			\copainstanceshort{The girl was not lonely anymore. Cause?}{She made a new friend.}
		}{
			\ratedexplanation{1.2}{Making is motivated by loneliness.}
		}

		\bottomrule
\end{tabular}

\caption{Examples of top-rated and bottom-rated explanations. Highly rated explanations tend to be detailed and explicitly connect the question and answer. Low rated ones are incoherent, completely irrelevant, or related facts but irrelevant as an explanation.}
\label{tab:extreme-examples}
\end{table*}

\renewcommand{\arraystretch}{0.95}

\subsection{Rating Explanations}
\label{sec:rating}

Figure \ref{fig:rating} shows our form for rating explanations.
Each explanation was rated by five workers.
Workers were shown a \copa{} instance and five explanations to rate with up to five stars. 
As a control, workers had to rate the first explanation again at the end of the HIT, totaling six ratings per HIT. We disregarded (but did not reject) ratings by workers who had more than a one-star difference in this control.\footnote{We  
  allowed a 1-star difference as one could change their opinion on the first seen explanation after seeing other examples. In case of such a difference, we only retain the last rating.}
Workers were instructed to give a higher rating to explanations containing relevant and more detailed statements and low ratings to uninformative or nonsensical explanations.
We observed that detailed, related statements were also low-rated if they did not explain why the answer is correct.
Examples of high-rated and low-rated explanations are shown in Table~\ref{tab:extreme-examples}.
While these ratings serve as generic estimate of quality, we recommend against using them as measurements of any single characteristic such as relevance or thoroughness since they were not defined as such.

\subsection{Re-collection}
\label{sec:recollecting}
To increase the number of higher-rated explanations, we invited workers who provided high-quality explanations to provide additional explanations for a higher fee.
We collected four new explanations for questions that had all five explanations rated below $3.5$-stars, two new explanations if one was above this threshold, and one new explanation if two were above this threshold. 
New explanations were then rated in the same way as the original ones.

\subsection{Compensation and qualifications}
\label{sec:qualifs}
Workers received $\$0.30$ per explanation in the first collection round and $\$0.40$ in the re-collection round. 
In the rating rounds, workers received $\$0.30$ for six ratings (five unique and one control).
We restricted all our rounds to workers in GB or the US with a HIT approval rate of $98\%$ or more and 500 or more approved HITs.
For re-collection, we invited workers whose explanations averaged more than $3.5$ stars over ten or more explanations. 
The total cost, including Amazon Mechanical Turk fees and excluding trial runs, was $\$8,651.16$.

\subsection{Post-processing: Aggregation}
\label{sec:aggregation}
Free-form nodes occasionally contain very similar concepts expressed with different surface forms without being explicitly connected.
Multiple explanations may also offer diverse information which, combined, results in a higher-quality explanation graph in terms of coverage.
To aggregate the explanations, we scored the similarity between each node and merged similar nodes or connected them with a \texttt{RelatedTo} edge. 
Specifically, we computed the cosine similarity $s$ of the node texts using Sentence-BERT~\cite{reimers-2019-sentence-bert} and merged if $s > 0.85$ or connected if $0.60 > s \geq 0.85$. 
The thresholds were manually determined by the authors with respect to the scores and resulting graphs.\footnote{For example, \textit{``sun''} and \textit{``under the sun''} are connected ($s$ = $0.76$), \textit{``shadow''} and \textit{``shadows''} are merged ($s$ = $0.93$). }
Each edge also includes a weight calculated as the sum of average human ratings of the explanation the edge came from.
Intuitively, these can be considered as the importance or relevance of the edge according to humans, at least in relation to all other given explanations for the sample.
Post-processed versions of the graphs are also available in the repository.

\section{Experiment}
\label{sec:experiments}
One use case of \ours{} is the creation of systems that automatically score explanations.
To demonstrate, we present baseline results on the task of outputting a quality rating given an explanation and, optionally, the question and the correct answer as additional context.
We evaluated the performance by measuring the Pearson correlation coefficient with human ratings and compared fine-tuned T5 \cite{roberts2019t5} implementations of various sizes ranging from 60M to 11B parameters.
Each was tested with the following input format: ``\textit{Rate this explanation:} \{premise\} \textit{so/because} \{correct\_answer\} \textit{Explanation:} \{explanation\}'' if including the QA context, and ``\textit{Rate this explanation: } \{explanation\}'' otherwise.
\footnote{\copa{} asks for the cause or the effect of a premise. An example input is as follows: ``Rate this explanation: \textit{My body cast a shadow over the grass.} because \textit{The sun was rising.} Explanation: \textit{Sunrise causes casted shadows.''} The gold rating for this explanation is $3.6$ (out of 5).}
In both cases, the expected output is the rating as a decimal number.
We followed the original \bcopa{} split and used the explanations for the original development questions, setting aside 5\% for validation, as training data, and explanations for the test questions as test data.

\begin{table*}
  \centering
  \begin{tabular}{lrcc}
      Model & Size & Expl. only & Expl.+QA context \\
      \toprule
      T5-Small  & 60M & 0.322 & 0.190 \\
      T5-Base & 220M & 0.476 & 0.504 \\
      T5-Large & 770M & 0.535 & 0.556 \\
      T5-3B & 3B & 0.530 & 0.569 \\
      T5-11B & 11B & 0.515 & \textbf{0.576} \\
      \bottomrule
  \end{tabular}
  \caption{Pearson's correlation between human ratings of \ours{} explanations and ratings outputted by fine-tuned T5 implementations of various sizes (num. parameters) given the same explanations. ``Expl.+QA context'' are the results when the models received both the explanation and the QA context as the input, and ``Expl. only'' when the input only included explanations. All values are averages over three runs.} \label{tab:t5-results}
\end{table*}

The results are shown in Table~\ref{tab:t5-results}.
Each value is the average Pearson coefficient over three runs.
Overall, correlation with human ratings increased with the size of the model and was generally higher when providing the QA context. 
However, even the best performing setting only reached a moderate correlation of $0.58$. 
This shows the potential of future explanation scoring systems trained with human-scored explanation data, with still much room for improvement.

\section{Discussion}
\label{sec:discussion}
\paragraph{Outlook: \ours{} as a Resource for Commonsense Reasoners.}
\label{sec:future-work}
We created this dataset with several uses in mind: it can serve as training data for (textual) explanation generation models, or as representations of ``ideal'' subgraphs to use as gold data for graph-based reasoners or to compare with existing KGs. 
\ours's textual explanations can be used to improve language model (LM)-based systems such as Commonsense Auto-Generated
Explanations (CAGE) \cite{rajani2019explain}, the system \cose{} was first intended for, which uses a LM to generate explanations as an intermediate step during training and inference.
Its triple-like format can in turn be useful for improving graph-based reasoning systems such as QA-GNN \cite{yasunaga2021qagnn}.
While still outperformed by current state-of-the-art systems, graph-based systems have the benefit of being more interpretable than purely LM-based systems due to having an accessible internal reasoning structure.
At the time of writing, the top-performing graph-based system is QA-GNN \cite{yasunaga2021qagnn}, a system combining LMs and GNNs to extract and weigh relevant knowledge from a KG, then perform reasoning over the extracted subgraph.
Our aggregated graph-form explanations (\S \ref{sec:aggregation}) can be considered as idealized versions of relevant subgraphs, thus offering gold examples for improving the extraction and relevance scoring steps in such a system.
Even though the explanations are in a similar format, their degree of freedom made it possible to collect new information that might not have been present in \cpnet{}.
We intend to further explore this direction with the primary goal of steering graph-based systems into being more interpretable.

\section{Conclusion}
We introduced a new crowdsourced dataset of explanations for \bcopa{} in a triple-based format intended for advancing graph-based QA systems and clear comparison with existing commonsense KGs.
The dataset provides relevant and minimal information needed to bridge the question and answer.
Our dataset includes explanations in text form and raw triple form as written by crowdworkers, and post-processed versions with similar nodes being merged or connected.
This dataset can serve to improve explanation generation in text-based or graph-based approaches.

\section{Acknowledgements}
This work was partially supported by JST CREST Grant Number JPMJCR20D2 and JSPS KAKENHI Grant Number 21K17814. 
Special thanks to Tatsuki Kuribayashi for his valuable advice.



\section{Bibliographical References}\label{reference}
\label{main:ref}

\bibliographystyle{lrec2022-bib}
\bibliography{bibliography}

\section{Language Resource References}
\label{lr:ref}
\bibliographystylelanguageresource{lrec2022-bib}
\bibliographylanguageresource{languageresource}

\end{document}